\documentclass[3p,12pt]{elsarticle}

\usepackage{amssymb}
\usepackage{amsmath}
\usepackage{array}
\usepackage{float}
\usepackage{url}
\usepackage{ulem}
\usepackage{amsmath}

\usepackage{xcolor}
\definecolor{xgreen}{HTML}{089b00}
\definecolor{xyellow}{HTML}{d9d92e}
\definecolor{xorange}{HTML}{ff7700}
\definecolor{xred}{HTML}{bb0000}
\definecolor{xblue}{HTML}{363e8e}

\newcolumntype{M}[1]{>{\centering\arraybackslash}m{#1}}
\usepackage{newtxmath}

\begin{document}

\begin{frontmatter}



\title{Benchmarking human-robot collaborative assembly tasks}


\author[inst1]{Laura Duarte$^*$}
\author[inst1]{Miguel Neves$^*$}
\author[inst1,*]{Pedro Neto}

\affiliation[inst1]{organization={Centre for Mechanical Engineering, Materials and Processes (CEMMPRE), ARISE},
            addressline={University of Coimbra},
            postcode={3030-788}, 
            state={Coimbra},
            country={Portugal}}
\affiliation[*]{Corresponding author: Pedro Neto, pedro.neto@dem.uc.pt}
\def\thefootnote{*}\footnotetext{These authors contributed equally to this work}

\begin{abstract}

Manufacturing assembly tasks can vary in complexity and level of automation. Yet, achieving full automation can be challenging and inefficient, particularly due to the complexity of certain assembly operations. Human-robot collaborative work, leveraging the strengths of human labor alongside the capabilities of robots, can be a solution for enhancing efficiency. This paper introduces the CT benchmark, a benchmark and model set designed to facilitate the testing and evaluation of human-robot collaborative assembly scenarios. It was designed to compare manual and automatic processes using metrics such as the assembly time and human workload. The components of the model set can be assembled through the most common assembly tasks, each with varying levels of difficulty. The CT benchmark was designed with a focus on its applicability in human-robot collaborative environments, with the aim of ensuring the reproducibility and replicability of experiments. Experiments were carried out to assess assembly performance in three different setups (manual, automatic and collaborative), measuring metrics related to the assembly time and the workload on human operators. The results suggest that the collaborative approach takes longer than the fully manual assembly, with an increase of 70.8\%. However, users reported a lower overall workload, as well as reduced mental demand, physical demand, and effort according to the NASA-TLX questionnaire.

\end{abstract}


\end{frontmatter}


\section{Introduction}
\label{sec:1}

Assembly processes play a pivotal role in the manufacturing industry, managing the growing complexity of product designs with an increasing number of components \cite{Vaclav2020}. They often integrate robots to enhance production efficiency, typically performing simple and repetitive operations. However, in the last decades, the production paradigm has shifted from mass production to customization in small to medium batch production. In such a scenario, traditional robots offer limited flexibility to the assembly processes as they are challenging to program, operate inside fences, and have relatively long setup times. The introduction of collaborative robots in manufacturing assembly leverages the best abilities of both robots and humans. When effectively combined, the robots versatility and the human’s flexibility result in a more productive and less error-prone workflow compared to fully manual or automated assembly processes \cite{Neto2019,ZHANG2024102691}. Nevertheless, defining collaborative scenarios remains challenging and highly dependent on the application itself, requiring the analysis of cost-benefit, safety issues, and the co-worker satisfaction \cite{Heo2023}. The technical specification ISO/TS 15066 establishes safety requirements for human-robot collaborative applications, encompassing limitations on the maximum robot velocity, a factor that significantly impacts the assembly time \cite{Saenz2020}.

Benchmarks are necessary to support users in making informed decisions regarding the adoption and deployment of human-robot collaborative systems in real working environments. They contribute to efficient and standardised research, enhancing reproducibility and facilitating result comparisons. The field of human-robot collaborative assembly currently lacks benchmarks encompassing multiple tasks with varying assembly complexities. Literature shows benchmarking setups featuring diverse experimental conditions and performance metrics with changing hardware and multiple methodologies for assessing the unpredictable human behaviour. Nevertheless, the choice between manual, automatic, or human-robot collaborative work, along with an understanding of the associated workload on humans remains unclear being highly dependent on the application.

\begin{figure*}[ht]
    \centering\includegraphics[width=1\textwidth]{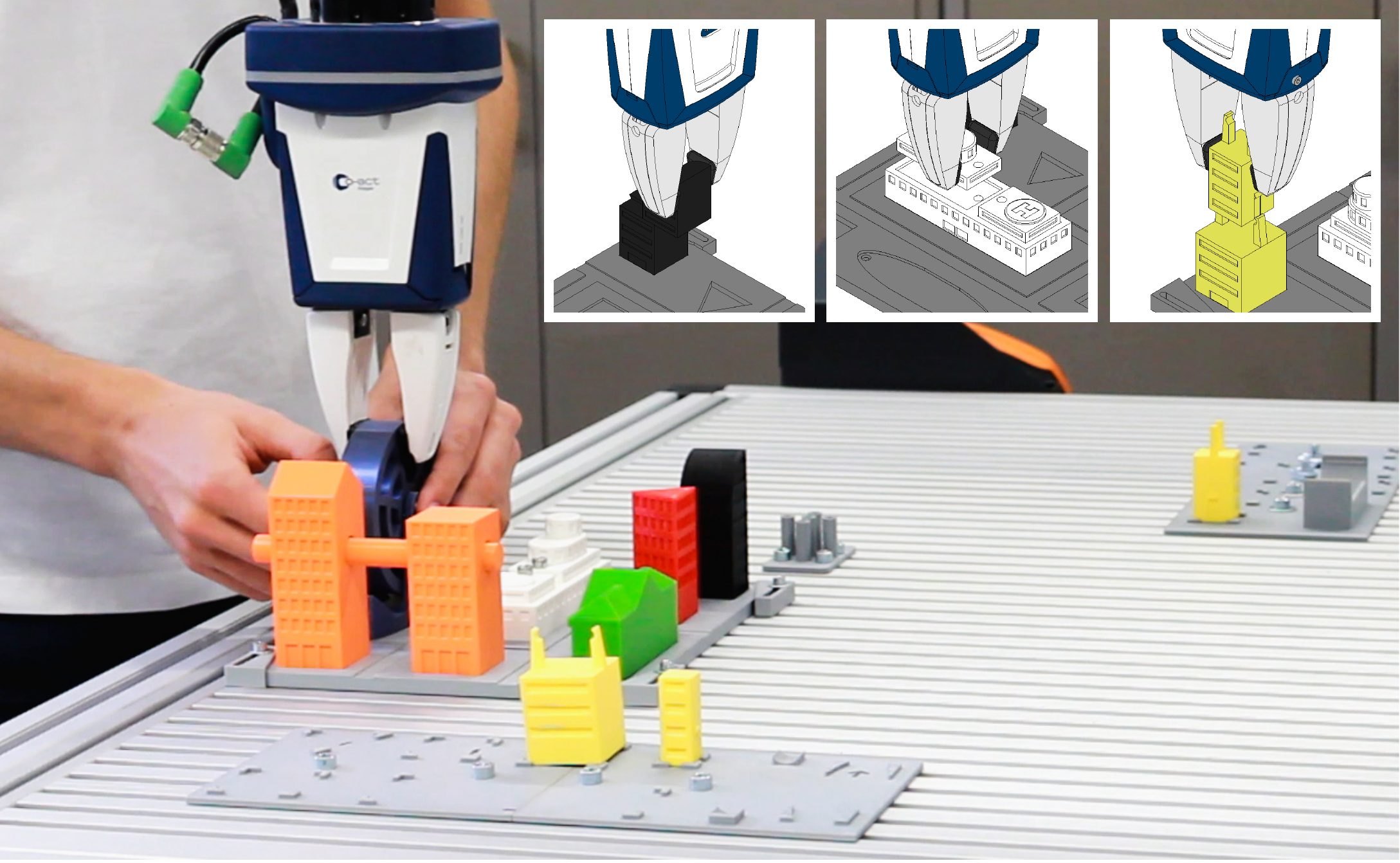}
    \caption{Human and robot collaborate to complete the assembly of the city landscape model proposed in the benchmark.}
    \label{fig:Setup}
\end{figure*}

In this paper, we propose a benchmark that involves assembling a city landscape model, covering a range of assembly scenarios with different levels of complexity, Fig.~\ref{fig:Setup}. The final assembly can be achieved by combining five common manufacturing assembly tasks. The benchmark comprises seven different sub-assemblies related to various buildings, each involving the assembly of one to three parts. The design of the benchmark was grounded to include independent sequential assembly tasks with a wide variety of sturdy and graspable 3D printable parts, each featuring distinguishable features. While the primary focus of the CT benchmark is to aid in the development of human-robot collaborative systems, it can also be applicable to task sequencing allocation, fully manual or automatic assembly, robot pick-and-place operations, and visual perception. The metrics considered are the total assembly time and the workload on human operators, assessed using the NASA-TLX questionnaire. The main contributions of this work are the following:

\begin{itemize}

\item A benchmarking framework specifically designed to evaluate and compare the performance of human-robot collaborative assembly tasks with fully manual and fully automatic assembly processes;
\item A rich set of components to be assembled through distinct tasks with varying levels of difficulty, inspired by common industrial assembly applications;
\item Open access to the STL models of the object set. These models were designed to be easily fabricated using a desktop 3D printer;
\item Fully manual, fully automatic, and collaborative assembly approaches are evaluated using quantitative metrics related to the assembly time and qualitative metrics in terms of the human workload from NASA-TLX questionnaire.
  
\end{itemize}

\section{Related work} 
\label{sec:2}

One of the earliest robot-related assembly benchmarks is commonly known as the Cranfield benchmark dedicated to pick-and-place and insertion tasks \cite{Collins1985}. It is composed of 5 parts and multiple spacers and securing pegs. Due to the limited number of different tasks in the Cranfield benchmark, Heemskerk and Luiten proposed the Delft Assembly Test (DATE) \cite{Heemskerk1988}. The primary design specifications for DATE contemplated the use of sensors, ensuring that the tasks could be achieved by any robot model, and involving orientation changes (vertical, 45$^{\circ}$, and horizontal assembly). Recently, various studies have proposed diverse metrics and benchmarks to evaluate human-robot interactive applications \cite{ALY2017313,9223347}. A benchmark for robotic assembly has been proposed using Assembly Task Boards \cite{Kimble2020}. It features a wide variety of insertion and fastening operations, such as threading, snap fitting, and gear meshing. Although it represents dexterous real-world assembly tasks, fully deploying the benchmark requires difficult-to-acquire components.

One common avenue for the design of benchmarks is the adoption of modular systems. Even though such benchmarks have the advantage of scalability, modular assemblies have inherent shortcomings. Tasks tend to be similar to each other, and different configurations make it difficult to make a fair comparison between research efforts. A benchmark for collaborative assembly of furniture through a modular and extendable design was introduced in \cite{Zeylikman2018}. This benchmark is cost-effective to deploy, and its constituent parts are not only readily available but can also be reused across a wide range of experiments. While it addresses the complexity and skill requirements of typical human-robot collaborative work, it lacks distinct assembly operations. On the other hand, the FurnitureBench is a recent benchmark specifically tailored for robotic manipulation in furniture assembly, focusing on long-horizon planning, dexterous control, and robust visual perception \cite{heo2023furniturebench}. Another modular benchmark emphasizes the concept of modular task boards where benchmark\textsc{\char13}s parts are grouped into three domains: cubes, conductors and gears \cite{Riedelbauch2022}. The main metrics for the benchmark\textsc{\char13}s design were relevance, representativeness, cost-effectiveness and time efficiency. A notable feature of this benchmark is the incorporation of real electrical circuitry into the 3D printed models. 

Benchmarks have also been introduced for various other robotics sub-areas and applications. The RLBench is a benchmark and learning environment designed for robot learning \cite{James2020}. It includes a variety of unique hand-designed tasks and it was built upon the virtual robot experimentation platform (V-REP) \cite{Rohmer2013}. In order to continually advance and improve the benchmark, task creation tools were developed. Another benchmark is presented for assessing mobile robot local planning approaches \cite{Wen2021}, encompassing diverse simulation scenarios and metrics such as safety, efficiency, and smoothness of motion. There are several benchmarks dedicated to robotic grasping and manipulation, with one of the most commonly used being the Yale-CMU-Berkeley Object and Model set \cite{Calli2015}. It covers a wide range of manipulation problems by incorporating a vast array of everyday objects with distinct physical properties. In it, several example protocols, each yielding a benchmark derived from it, are developed to quantify performance.

\begin{figure*}[ht]
	\centering\includegraphics[width=\textwidth]{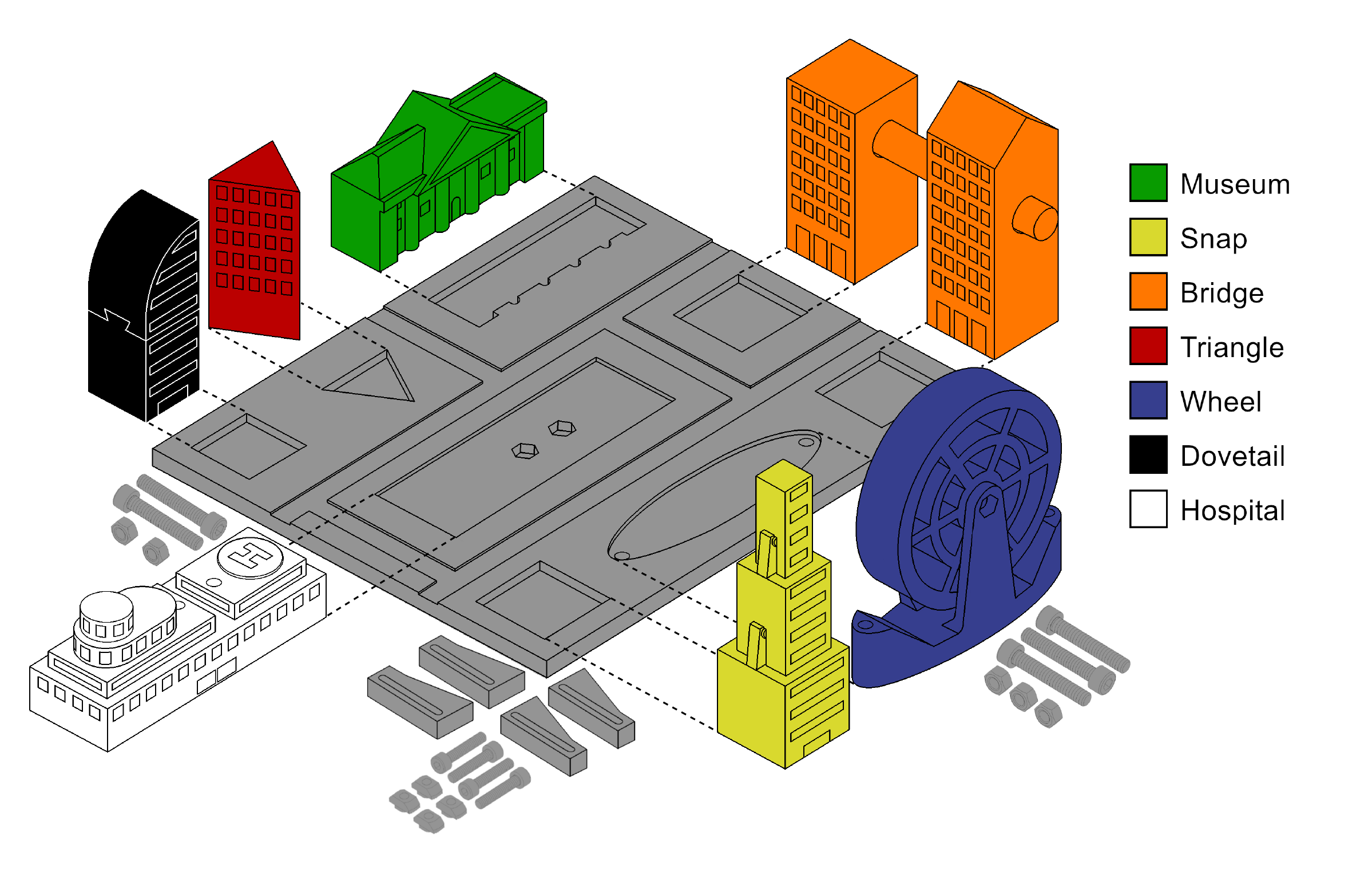}
	\caption{The object set of the CT benchmark, depicting a city landscape model combining five common manufacturing tasks which can be performed by humans or robots. The benchmark comprises seven different sub-assemblies related to the various buildings.}
	\label{fig:Setup2}
\end{figure*}

\section{CT benchmark }
\label{sec:3}

The CT benchmark is dedicated to the assembly of a city landscape model, covering a range of assembly scenarios with different levels of complexity, Fig.~\ref{fig:Setup2}. The benchmark comprises seven different sub-assemblies related to various buildings, represented by different colours, each involving the assembly of one to three parts, Table~\ref{table:Table}. The final assembly can be achieved by combining five common manufacturing assembly tasks, namely the wide tolerance insertions (WTI), the tight tolerance insertion (TTI), screw fastening, snap fitting and two-handed actions. The design of the benchmark contemplated independent sequential assembly tasks with a wide variety of sturdy and graspable 3D printable parts, each featuring distinguishable features. These tasks can be performed by a human, a robot, or both in a human-robot collaborative system.

\begin{table}[ht!]
\renewcommand{\arraystretch}{1.3} 
\centering
\caption{Benchmark sub-assemblies and related tasks.}
\begin{tabular}{c c c c | c c c c c}
 \\ [-10pt] 
 \hline
 Sub-assembly & Color & \# sub-tasks & \# parts & WTI & TTI & Screw & Snap-fit & Two-handed \\ 
 \hline
 Museum & \fcolorbox{black}{xgreen}{\rule{0pt}{6pt}\rule{6pt}{0pt}} & 1 & 1 & \checkmark & & & & \\
 Snap & \fcolorbox{black}{xyellow}{\rule{0pt}{6pt}\rule{6pt}{0pt}} & 3 & 3 & \checkmark & \checkmark & & \checkmark & \\
 Bridge & \fcolorbox{black}{xorange}{\rule{0pt}{6pt}\rule{6pt}{0pt}} & 3 & 3 & \checkmark & \checkmark & & & \\
 Triangle & \fcolorbox{black}{xred}{\rule{0pt}{6pt}\rule{6pt}{0pt}} & 1 & 1 & \checkmark & & & & \\
 Wheel & \fcolorbox{black}{xblue}{\rule{0pt}{6pt}\rule{6pt}{0pt}} & 3 & 2 & \checkmark & & \checkmark & & \checkmark  \\
 Dovetail & \fcolorbox{black}{black}{\rule{0pt}{6pt}\rule{6pt}{0pt}} & 2 & 2 & \checkmark & \checkmark & & & \\
 Hospital & \fcolorbox{black}{white}{\rule{0pt}{6pt}\rule{6pt}{0pt}} & 4 & 3 & \checkmark &  & \checkmark & & \\ 
 \hline
\end{tabular}
\label{table:Table}
\end{table}

\subsection{Objects}
\label{sec:3.1}
The benchmark\textsc{\char13}s STL files for the base, buildings and corner fixing pieces are openly available online\footnote{\url{https://github.com/Robotics-and-AI/collaborative-tasks-benchmark}}. For a wider dissemination and availability of the CT benchmark, most parts can be 3D printed with any commercially available 3D printer. The full deployment of the benchmark also requires five M5x30 screws and five ISO 4032 M5 nuts. Additionally, if the user intends to fix the base to a T-slot aluminium extrusion frame, the corner fixing pieces need to be 3D printed and four M4x20 screws and four adequately sized M4 T-nuts are required. Some of the buildings can be placed in multiple orientations, for example, the Dovetail, Snap and Triangle buildings. The benchmark parts were designed to be graspable, by humans and a robot gripper, requiring a grasping range between 14 mm and 40 mm.

Printing the entire benchmark requires $\sim540$g of material ($\sim520$g for the benchmark and $\sim20$g for the corner pieces) and takes $\sim26$ hours ($\sim25$ hours for the benchmark and $\sim1$ hour for the corner pieces). This information was gathered using the Prusa Slicer software with a 1.75 mm PLA filament, a 0.3 mm layer height and a 15\% grid infill. In our deployment of the benchmark, distinct colours were used for each sub-assembly, but the benchmark can also be printed in a single colour.

\subsection{Object assembly}

The Museum and Triangle sub-assemblies are the most elementary ones since they only require a single WTI task for each of the buildings into their respective allotted space to be complete. The Museum has the most complex base shape of the benchmark making the WTI task more challenging. The Bridge and the Dovetail are slightly more complex to assemble as they are comprised of multiple tasks and require precise placement for the TTI task. To fully assemble the Bridge, first, the two vertical buildings have to be inserted in their designated slots (WTI task) to then be able to insert the cylindrical part. The Dovetail has two tasks where the bottom part of the building must first be inserted in its designated spot, followed by the assembly of the top part through the Dovetail joint. The Snap building requires three tasks to be assembled. First, the bottom part is inserted in its location, then the middle part is assembled through the snap-fit connection and, finally, the top part is assembled also through a snap-fit connection. 

The Wheel and Hospital require the usage of screws and nuts to fully complete the assembly. The Wheel sub-assembly has a total of three tasks. First, its base must be inserted in the predefined slot, then two screws are used to affix it to the city base, and lastly, the wheel must be held in the correct position to fasten the last screw. Since the wheel must be held in place, two hands are required to finish the assembly. Finally, the Hospital has four tasks where its bottom part has to be inserted in its position, after which the top parts are both inserted in their respective slots (two tasks). To complete the task, the top and bottom hospital parts are subsequently fastened together, with one screw for each of the top parts.

\subsection{Protocol}
The benchmark objects should be placed on a horizontal surface within an area easily accessible to both humans and robots. They can be placed at known poses (positions and orientations) to facilitate robot grasping without the need for perception systems. The final assembly can be achieved using different sequences, and humans may use a screwdriver in the process.

\begin{figure*}[ht] 
	\centering\includegraphics[width=0.9\textwidth]{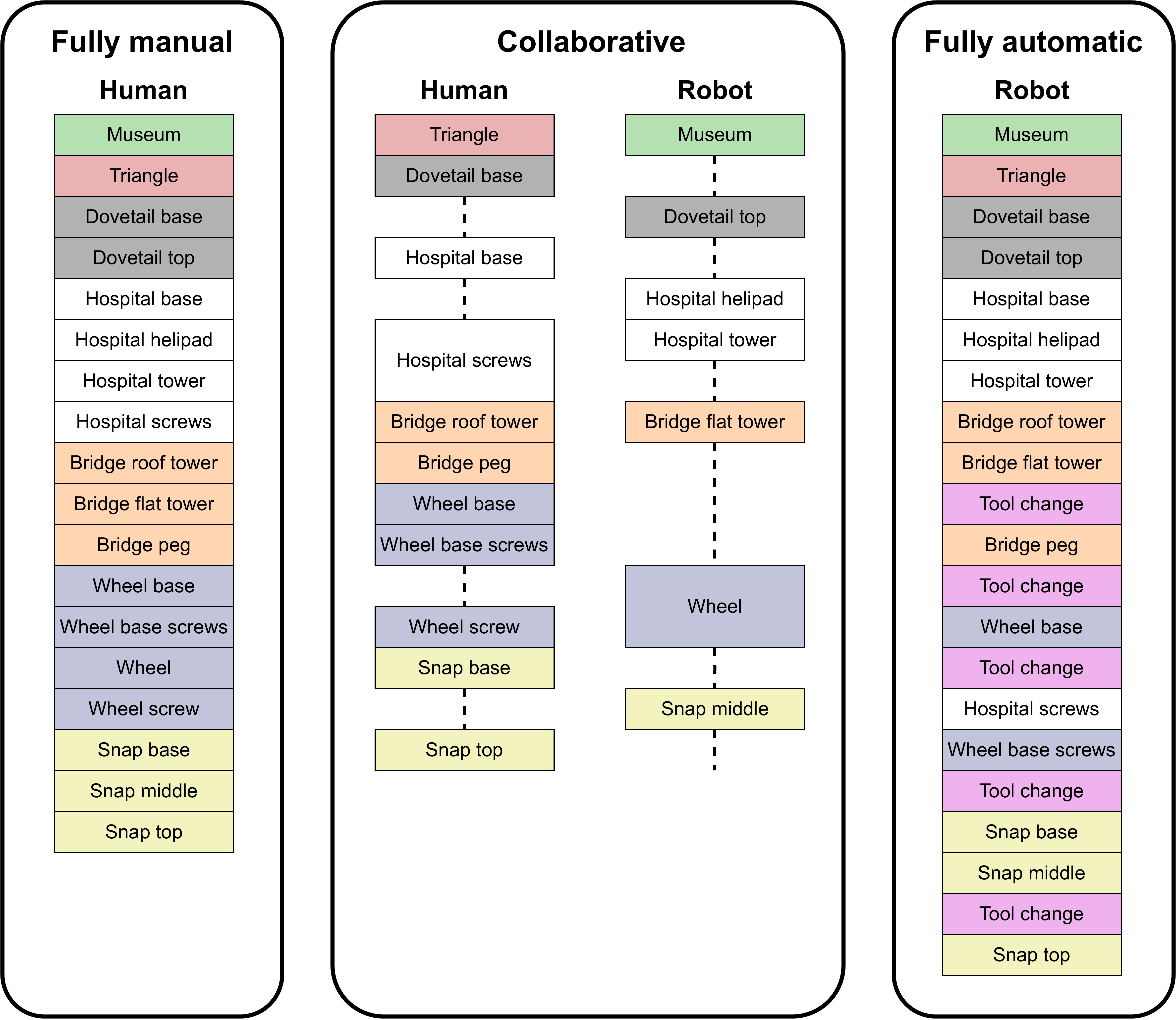}
		\caption{Schematic of a proposed benchmark sub-assembly sequencing for each setup: fully manual, collaborative and fully automatic. Tasks related to each sub-assembly are detailed in Table~\ref{table:Table}.}
	\label{fig:Sequence}
\end{figure*}

The benchmark was designed to promote the usage of robots with two distinct tools, a gripper and a screwdriver. In this work, we propose three illustrative setups, a fully manual assembly, a human-robot collaborative assembly and a fully automatic assembly process performed by a robot. The assembly sequence for these setups is similar to assure a fair comparison between them, Fig.~\ref{fig:Sequence}. In the fully automatic setup, the robot has to change the tools five times.

In a human-robot collaborative assembly, the benchmark can be tackled in a multitude of ways. Each task may be assigned either to the human, to the robot or even be assigned to the first element available. For example, the Bridge sub-assembly has three tasks, allowing for eight different ways to assign them, out of which six can be collaborative tasks, one a fully manual task and another one a fully automatic task performed by a robot.

\begin{figure*}[ht] 
	\centering\includegraphics[width=\textwidth]{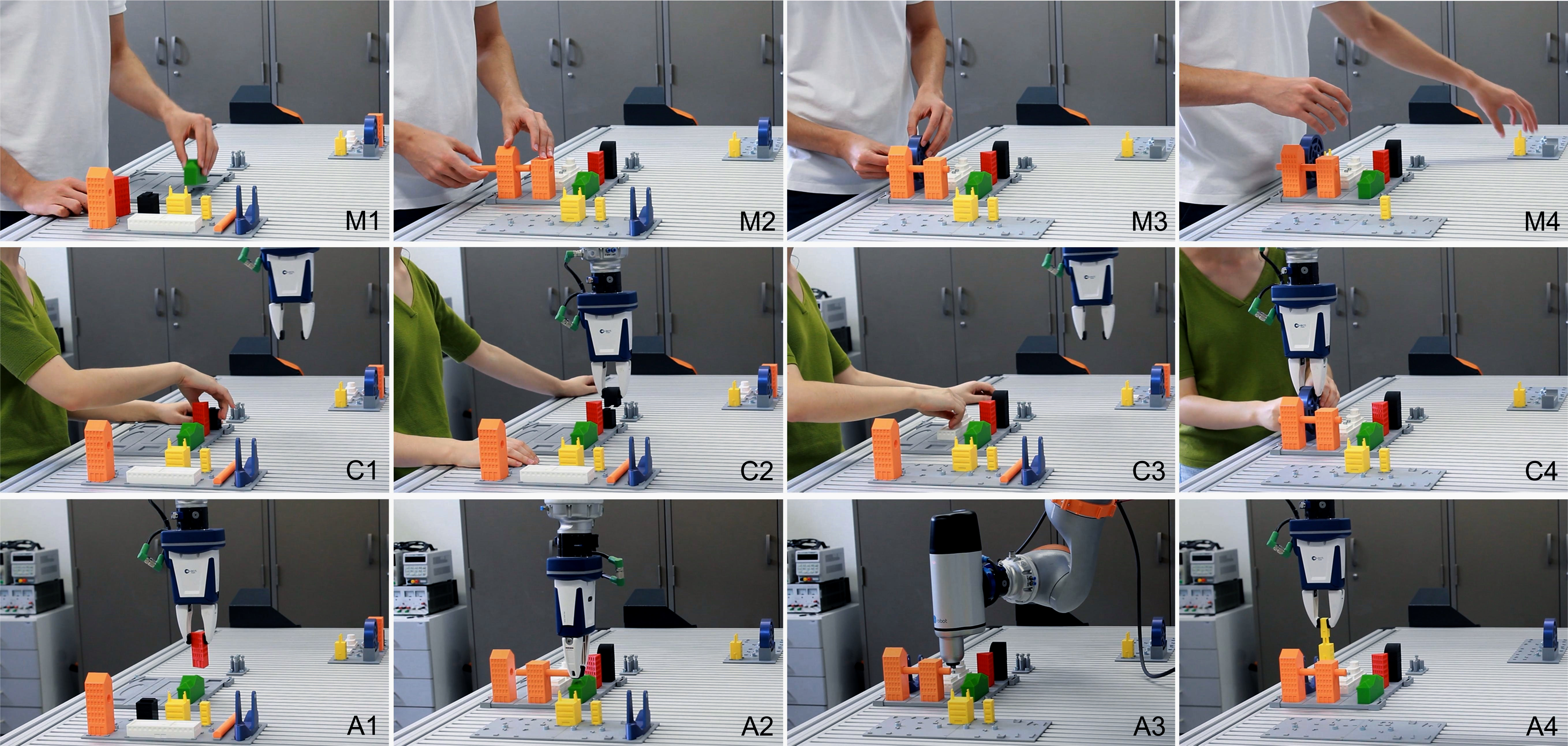}
	\caption{Snapshots of the proposed benchmark being used in fully manual assembly (M1 to M4), human-robot collaborative assembly (C1 to C4), and fully automatic assembly performed by a robot (A1 to A4). In the collaborative setup, the robot is equipped with a gripper, while in the fully automatic setup, the robot is equipped with either a gripper or a screwdriver tool.}
	\label{fig:Assemblies}
\end{figure*}

Examples of each assembly approach can be observed in Fig.~\ref{fig:Assemblies}. In the manual assembly, the human operator performs a WTI task (M1), a TTI task (M2), and a screw fastening task together with a two-handed task (M3). In the collaborative scenario, the human operator is performing a WTI task (C1), and screwing tasks without and with robot assistance, respectively (C3 and C4). The operator is idle while the robot performs a TTI task (C2). In the automatic mode, the robot can be observed performing a WTI task (A1), a TTI task (A2), a screwing task (A3) and a snap-fit task (A4).

\subsection{Metrics and evaluation}

Real-world applications of robots in collaborative assembly processes necessitate metrics to compare the performance of various approaches and setups. Such comparisons can aid in decision-making related to defining robot and assembly parameters, evaluating assembly time, and comparing between fully manual, collaborative, or fully automatic processes. These considerations should be adapted according to practical user needs.

Three assembly setups (fully manual, collaborative and fully automatic) were compared using both qualitative and quantitative metrics. Qualitative metrics could not be applied to the fully automatic approach, as it does not involve a human user. The fully automatic process was executed only once, given its identical nature in each repetition. Both the collaborative and fully manual assembly processes were performed once by each participant. Before starting any task, participants received brief instructions on the part locations and how to interact with them. To minimize bias, half of the participant group performed the manual assembly first, while the other half performed the collaborative assembly first. A total of fifteen participants performed the experiment, from whom eight executed the manual assembly first and seven started with the collaborative assembly.

The evaluation metrics considered are the total assembly time, a main parameter in manufacturing assembly, and the workload on human operators assessed using the NASA-TLX questionnaire. For the fully automatic setup, since our robot setup does not allow for a quick change of tools, a rough approximation of tool change time is defined as 10 seconds. The setups involving human labour are qualitatively evaluated using a questionnaire to assess the operator workload, the NASA Task Load Index (NASA-TLX). It estimates the workload of a human operator through six subscales: Mental Demand, Physical Demand, Temporal Demand, Frustration, Effort, and Performance \cite{Hart2006}. A common modification often made to NASA-TLX, referred to as Raw TLX, is to eliminate its subscale weighting process and analyse them individually. It is given to the participants after completing both assembly scenarios, to identify the benefits and drawbacks of using a collaborative robot in comparison to a fully manual assembly. 

The collaborative robot is a KUKA iiwa collaborative equipped with a SCHUNK Co-act gripper with a grasping range between 20 mm and 40 mm. The fully automatic setup requires an additional robot equipped with a screwdriver tool. In our experiments, we use a single robot, and thus the two-handed task is missing from the wheel sub-assembly.

\section{Results and discussion}
\label{sec:4}

\subsection{Quantitative Analysis}
\label{sec:4.1}

\begin{figure*}[ht] 
    \centering\includegraphics[width=0.85\textwidth]{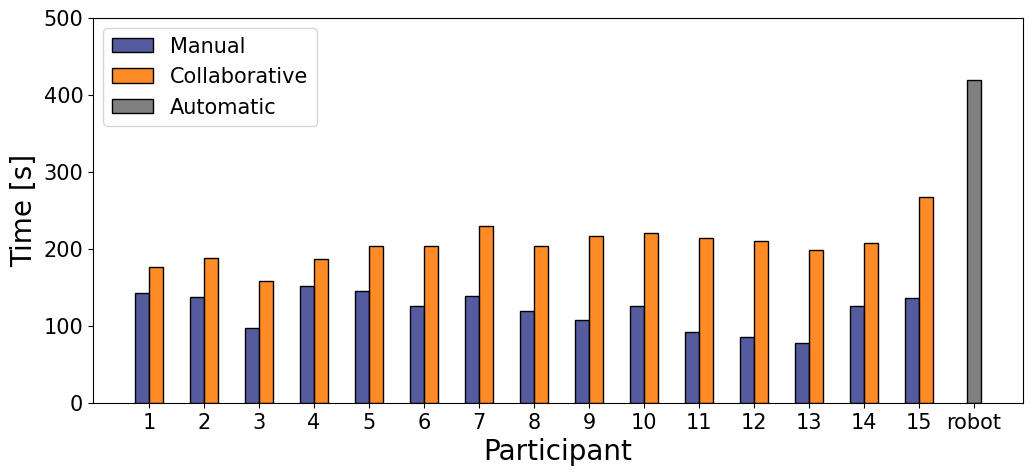}
    
    \caption{Total time registered for each participant after performing the manual and collaborative assemblies. Participants 1 to 8 first executed the manual assembly and participants 9 to 15 initially executed the collaborative assembly. The last result illustrates the total time registered for the fully robotic assembly.}
    \label{fig:Totaltimes}
\end{figure*}

The total assembly times registered for each of the setups are presented in Fig.~\ref{fig:Totaltimes}. A significant reduction can be seen in the total times of the manual assemblies from the participants who performed the collaborative assembly first. The opposite also holds true, yet is less distinctive. This is to be expected, as the user gets accustomed to the benchmark and thus performs better on the second assembly. Overall, the manual assemblies take less time, averaging about 120 seconds (2 minutes), while the collaborative assemblies take about 70.8\% longer, averaging 205 seconds (3 minutes and 25 seconds). The total assembly time recorded for the fully automatic assembly is 419 seconds (6 minutes 59 seconds), excluding tool changes. Five tool changes were performed, each increasing the assembly duration by 10 seconds, for a total of 469 seconds (7 minutes and 49 seconds), Fig.~\ref{fig:Sequence}. As such, the fully automatic assembly is the slowest approach. Even when defining fast robot movements between picking and placing an object, the insertions require fine motor control, to which a robot needs lower speeds to perform accurately and repeatedly.

\begin{figure*}[ht] 
    \centering\includegraphics[width=0.8\textwidth]{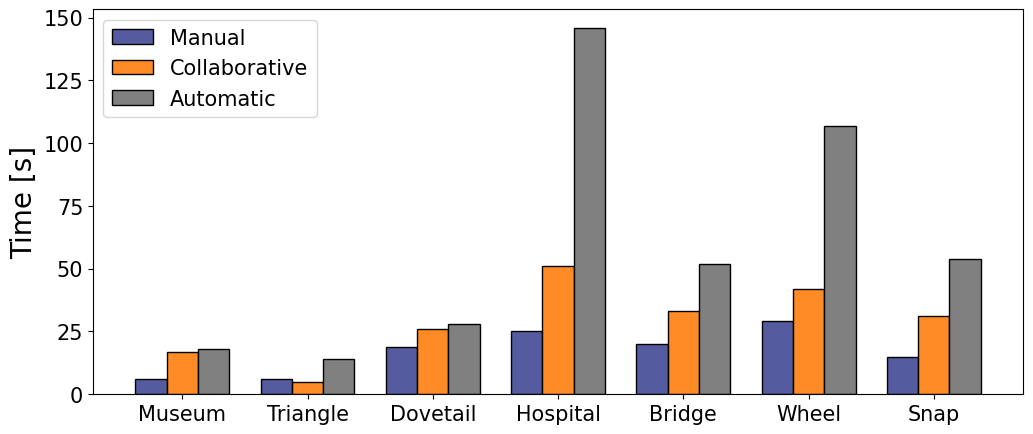}
    
    \caption{Average sub-assembly time for each setup. Average times recorded from the fifteen participants.}
    \label{fig:Tasktimes}
\end{figure*}

For both manual and collaborative assembly scenarios, the average time of each sub-assembly was recorded, Fig.~\ref{fig:Tasktimes}. In the case of the fully automatic assembly, screwing tasks were done back-to-back to require fewer tool changes while retaining a similar order of assembly to the other approaches, as depicted in Fig.~\ref{fig:Sequence}. Overall, collaborative tasks take longer to execute than manual tasks. This discrepancy arises because the human operator often has to wait for the robot to finish an operation before continuing with the assembly process. Additionally, the human operator is faster at retrieving the benchmark parts within arm's reach, being able to move in an agile and efficient manner and using both hands to retrieve multiple parts. In contrast, the robotic arm has speed limitations when working in collaborative scenarios due to safety regulations, and the insertion of parts must be done with slow, controlled movements. Alternatively, performing multiple assemblies in parallel could increase throughput within the same allotted time frame.

The fully automatic mode takes longer to perform than its collaborative counterparts, as the latter distributes the workload between the human operator and the robot. As expected, a time increase was observed from fully manual assembly to collaborative assembly, and from the collaborative to the fully automatic mode. A significant discrepancy between collaborative and automatic assemblies can be observed, particularly in two sub-assemblies: the Hospital and the Wheel. It is evident that the increase in duration is mainly due to the fact that both tasks feature two screws each. In our particular robot configuration, the robot had to be moved slowly to guarantee screw alignment, amplifying the time difference. Regardless, even when considering optimal conditions and dismissing time penalties due to hardware limitations, a collaborative assembly still outperforms the automatic assembly in terms of time efficiency.

\subsection{Qualitative Analysis}
\begin{figure*}[ht] 
    \centering\includegraphics[width=0.8\textwidth]{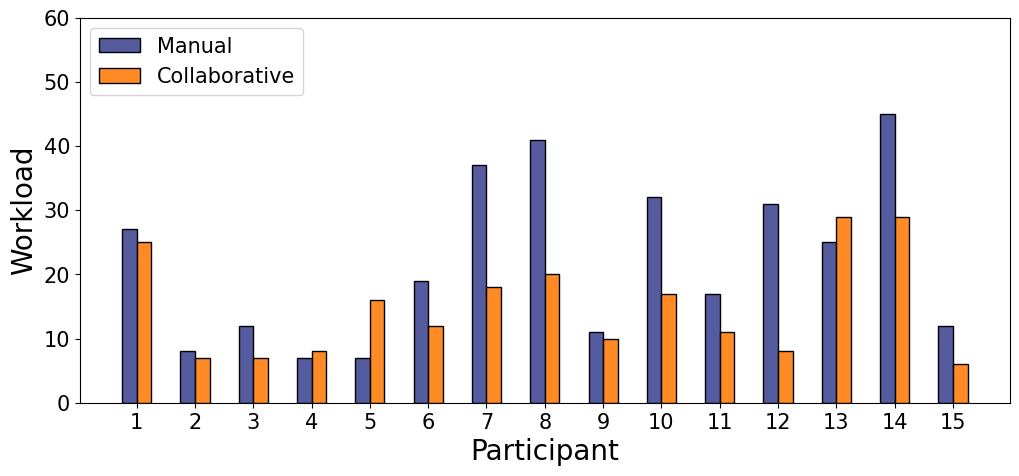}
    
    \caption{Workload reported by each participant after performing the manual and collaborative assemblies. Participants 1 to 8 executed the manual assembly first and participants 9 to 15 initially executed the collaborative assembly.}
    \label{fig:Workload}
\end{figure*}

Each participant answered a modified Raw TLX questionnaire after executing both assemblies. The only change made to the original Raw TLX was the reduction of the scale to 10 steps. From the questionnaire results the tendency to assign a lower workload to the collaborative assembly is clear, as reported by 12 out of the 15 participants, Fig.~\ref{fig:Workload}.

\begin{figure*}[ht] 
    \centering\includegraphics[width=0.9\textwidth]{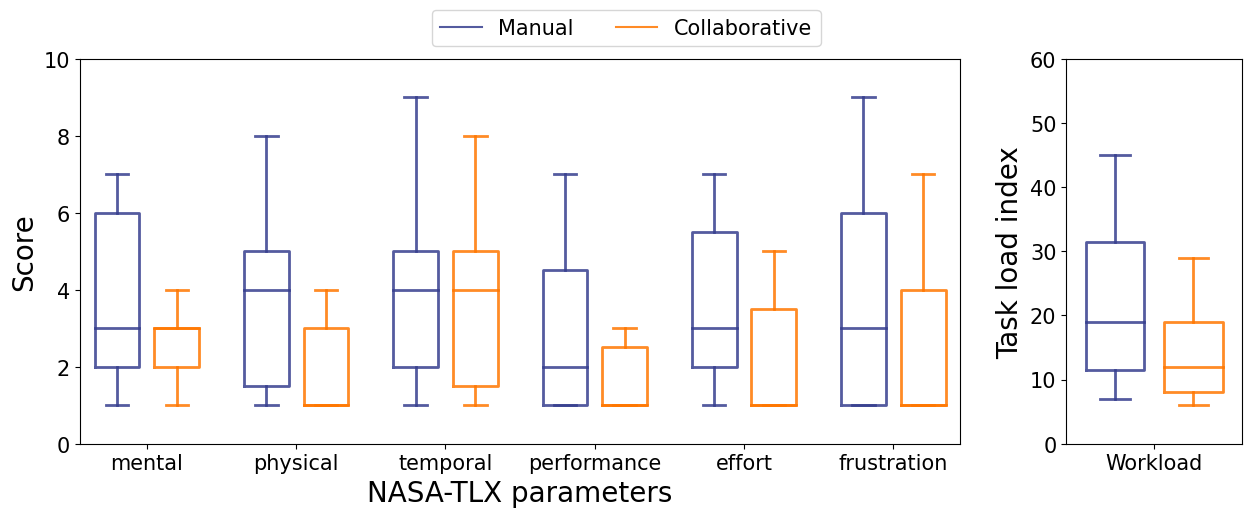}
    
    \caption{Questionnaire results for each parameter and total workload. Mental demand, physical demand, effort and workload have a statistically significant difference.}
    \label{fig:Boxplot}
\end{figure*}

Each participant answered the same questionnaire for both assemblies (manual and collaborative). As such, the samples are dependent.
However, the differences between manual and collaborative workloads are not normally distributed, as can be observed when plotting the corresponding histogram. As such, to verify statistical significance, the non-parametric Wilcoxon test is used. The Wilcoxon test was applied individually to each one of the NASA-TLX parameters and to the total workload with an alpha value of 0.05. The mental demand, physical demand and effort parameters, as well as the total workload, are found to be statistically different between the two approaches. As such, it can be concluded that the collaborative assembly has a significant reduction of the felt workload to the human operators. These results support the major differences observed in the results for the NASA-TLX questionnaire for both assemblies, Fig.~\ref{fig:Boxplot}.

\section{Conclusions}
\label{sec:6}

In this paper, we introduce the CT Benchmark and model set designed to evaluate human-robot collaborative assembly scenarios and compare them with manual and automatic processes using key performance metrics such as assembly time and human operator workload. The STL model set is provided in open access, allowing for straightforward deployment of the benchmark through 3D printing. The benchmark offers a wide range of assembly tasks, featuring insertion and fastening operations attributable to either a human operator or a robot, providing a rich environment for testing collaborative systems. We present some example evaluations comparing the performance of collaborative assembly setups with fully manual and fully automatic processes. Results demonstrate that the workload, as reported on the NASA-TLX questionnaire, is significantly reduced when using a collaborative approach, albeit at a cost of 70.8\% longer assembly times when compared to the fully manual assembly. The results obtained for mental demand, physical demand, and effort significantly improved with the collaborative approach.

\section*{Declaration of Competing Interest}
The authors declare that they have no known competing financial interest or personal relationships that could have appeared to influence the work reported in this paper.

\section*{Acknowledgments}
This research is sponsored by national funds through FCT – Fundação para a Ciência e a Tecnologia, under the project UIDB/00285/2020 and LA/P/0112/2020, and the grants 2021.06508.BD and 2021.08012.BD.

\bibliographystyle{plain} 




\end{document}